\documentclass[journal,twoside,web]{ieeecolor}
\usepackage{generic}
\usepackage{cite}
\usepackage{amsmath,amssymb,amsfonts}
\usepackage{algorithmic}
\usepackage{graphicx}
\usepackage{algorithm,algorithmic}
\usepackage{hyperref}
\hypersetup{hidelinks=true}
\usepackage{textcomp}
\usepackage{booktabs} 
\usepackage{multirow}
\usepackage{placeins}
\def\BibTeX{{\rm B\kern-.05em{\sc i\kern-.025em b}\kern-.08em
    T\kern-.1667em\lower.7ex\hbox{E}\kern-.125emX}}
\begin{document}
\title{MedVL-SAM2: A unified 3D medical vision-language model for multimodal reasoning and prompt-driven  segmentation}
\author{Yang Xing, Jiong Wu, Savas Ozdemir, Ying Zhang, Yang Yang, Wei Shao and Kuang Gong
\thanks{This work was supported by NIH grants R01EB034692 and R01AG078250 (Corresponding author: Kuang Gong).}
\thanks{X. Yang, J. Wu and K. Gong are with J. Crayton Pruitt Family Department of Biomedical Engineering, University of Florida, Gainesville, FL, USA (e-mail: yxing2@ufl.edu, wujiong@ufl.edu, kgong@bme.ufl.edu).}
\thanks{S. Ozdemir is with the Department of Radiology, University of Florida, Jacksonville, FL, USA (e-mail: savas.ozdemir@jax.ufl.edu).}
\thanks{Y. Zhang is with the Department of Research Computing, University of Florida, Gainesville, FL, USA (e-mail: yingz@ufl.edu).}
\thanks{Y. Yang is with the Department of Radiology, UC San Francisco, San Francisco, CA, USA (e-mail: Yang.Yang4@ucsf.edu).}
\thanks{W. Shao is with the Department of Medicine, University of Florida, Gainesville, FL, USA (e-mail: Wei.Shao@medicine.ufl.edu).}}

\maketitle

\begin{abstract}
Medical vision–language models (VLMs) have shown strong potential on image-level, text-centric tasks such as report generation and visual question answering (VQA). However, achieving fine-grained visual localization and volumetric spatial reasoning in 3D medical VLMs remains challenging, particularly when attempting to unify these capabilities within a single framework. To address this challenge, we propose MedVL-SAM2, a unified 3D medical VLM that simultaneously supports report generation, VQA, and multi-paradigm segmentation, including semantic, referring, and interactive segmentation. MedVL-SAM2 integrates image-level reasoning and pixel-level localization through a cohesive architecture tailored for 3D medical imaging. The model is trained via a multi-stage pipeline. First, it is pre-trained on a large-scale collection of 3D CT image–text pairs to align volumetric visual representations with radiology language embeddings. It is then jointly optimized using both language understanding and segmentation objectives on a comprehensive 3D CT segmentation dataset. This joint training enables flexible interaction through language, point, or box prompts, effectively unifying high-level semantic reasoning with spatially precise localization. Experiments demonstrate that MedVL-SAM2 achieves state-of-the-art performance across report generation, VQA, and diverse 3D segmentation tasks. These results suggest that high-level semantic reasoning and precise volumetric localization can be effectively integrated within a unified 3D medical VLM framework. Code is publicly available at https://github.com/astlian9/MedVL-SAM2. 
\end{abstract}

\begin{figure}[t]
  \centering
   \includegraphics[width=\linewidth]{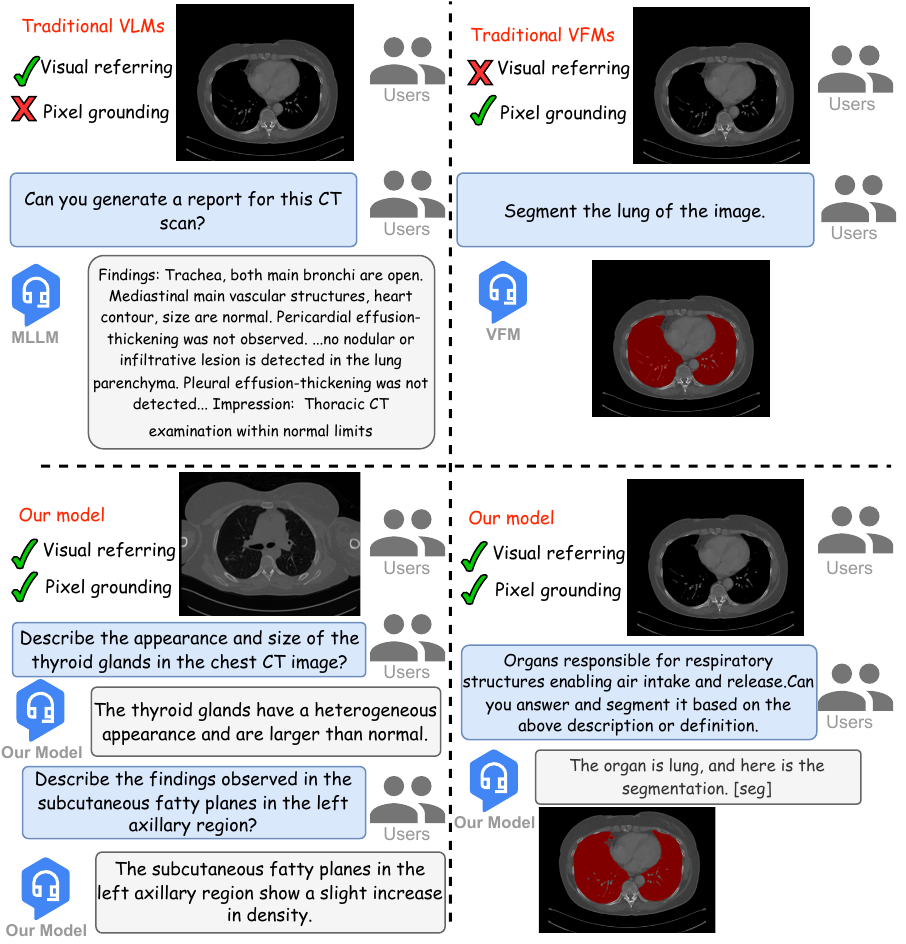}
   \caption{Comparison of the proposed model to traditional text-response-only medical vision-language models (VLMs) and mask-response-only medical vision foundation models (VFMs).  }
   \label{fig:onecol}
\end{figure}
\section{Introduction}
\label{sec:introduction}
\IEEEPARstart{L}{everaging} large-scale radiology image-text pairs, medical vision-language models (VLMs) exhibit strong reasoning ability over both global image semantics and clinical narratives. Furthermore, several 2D VLM studies such as MIMO~\cite{chen2024mimo} and MedPLIB~\cite{huang2024biomedpixelllm} have extended VLMs with spatial localization capabilities, enabling them to identify disease-related regions or anatomical structures referenced in language prompts. However, the reliance on 2D vision-language architectures inherently constrains their effectiveness in 3D medical image analysis. Specifically, 2D VLMs lack volumetric perception and spatial consistency across slices, both of which are essential for understanding complex 3D structures such as tumors, vessels, and organs. Consequently, while 2D VLMs may identify features within an individual slice, it fails to capture the spatial continuity, anatomical context, and morphological variations present across the full 3D volume.

Recent efforts have begun to explore 3D medical VLMs, such as Med3DVLM~\cite{xin2025med3dvlm} and CT-CHAT~\cite{hamamci2024generalistCT}, which process volumetric data and generate textual outputs. Nevertheless, these models remain limited to language-only output, lacking voxel-level localization or segmentation capabilities, which restricts their applicability in clinically meaningful interpretation and interaction. A significant gap thus remains in developing 3D medical VLMs capable of fine-grained spatial localization and aligning textual concepts with 3D spatial regions. To address this, several 3D VLMs with segmentation abilities have been proposed, including M3D~\cite{bai2024m3d} and VILA-M3~\cite{nath2024vilam3}. M3D incorporates 3D information but is constrained by the low input resolution of its vision encoder and the design of its projection layer, introducing substantial volumetric information loss. Meanwhile, VILA-M3 relies on domain-specific expert modules for downstream tasks, and its performance may become unstable when these components are not effectively activated. These limitations underscore the need for a unified 3D medical VLM capable of understanding volumetric image structures while preserving strong language-vision alignment.

To address these challenges, we propose MedVL-SAM2, a unified 3D medical VLM that extends a LLaVA-style architecture to incorporate volumetric perception and multimodal reasoning. Fig.~\ref{fig:onecol} illustrates the differences between our model and traditional medical VLMs and vision foundation models (VFMs). MedVL-SAM2 incorporates image-level understanding and pixel-level segmentation within a single framework, enabling both high-level language-driven analysis and fine-grained spatial localization. Specifically, our model integrates the prompt-driven paradigm of SAM2 into a 3D vision-language backbone, supporting multimodal interaction through language, point, and bounding-box prompts. Our key contributions are summarized as follows:
\begin{itemize}
    \item The proposed model provides a unified 3D medical VLM capable of jointly performing image-level reasoning and pixel-level perception, supporting report generation, VQA, and multiple types of 3D segmentation.
    \item The framework incorporates multimodal prompting through language, point, and bounding-box inputs, enabling flexible and interactive 3D image segmentation.
    \item The method achieves superior performance on two large-scale benchmarks, consistently surpassing current approaches in both reasoning and segmentation tasks.
\end{itemize}

\section{Related Works}
\label{sec:formatting}
\begin{table}
  \caption{Capability comparison between our model and current state-of-the-art methods across report generation (RG), visual question answering (VQA), semantic segmentation (SS), referring segmentation (RS), and interactive segmentation (IS).}
  \label{tab:task_comparison}
  \centering
  \tiny
  \renewcommand{\arraystretch}{1}
	\resizebox{0.9\columnwidth}{!}{%
  \begin{tabular}{@{}lccccc}
    \toprule
    Method & RG & VQA & SS & RS  & IS\\
    \midrule
    BiomedParse~\cite{zhao2025biomedparse}  & $\times$ & $\times$ & $\checkmark$  & $\checkmark$ & $\times$ \\
    medSAM2~\cite{ma2024segmentmedicalimagesvideos}  & $\times$ & $\times$ & $\checkmark$  & $\times$ & $\checkmark$ \\
    CT-CHAT~\cite{hamamci2024generalistCT} & $\checkmark$ & $\checkmark$ & $\times$  & $\times$ & $\times$ \\
    Med3DVLM~\cite{xin2025med3dvlm} & $\checkmark$ & $\checkmark$  & $\times$ & $\times$ & $\times$ \\
    Med-2E3~\cite{shi2025med2e32denhanced3dmedical} & $\checkmark$ & $\checkmark$ & $\times$  & $\times$ & $\times$ \\
    MS-VLM~\cite{lee2024readlikeradiologistefficient}  & $\checkmark$ & $\checkmark$ & $\times$ & $\times$ & $\times$ \\
    M3D~\cite{bai2024m3d}   & $\checkmark$ & $\checkmark$ & $\checkmark$ & $\checkmark$ & $\times$ \\
    Ours &  $\checkmark$ & $\checkmark$ & $\checkmark$ & $\checkmark$ & $\checkmark$ \\
    \bottomrule
  \end{tabular}
  }
\end{table}
\subsection{VLMs}
VLMs integrate visual encoders with large language models (LLMs) for vision-language understanding. Early approaches rely on complex fusion mechanisms, such as cross-attention (Flamingo~\cite{alayrac2022flamingo}) and query transformers (BLIP-2~\cite{li2023blip2}), while more recent approaches have converged toward simpler, unified designs. LLaVA~\cite{liu2023vit,li2024llavaonevisioneasyvisualtask} shows that a vision encoder with a linear projection is sufficient for effective vision-language alignment, leading to numerous extensions (LLaVA-Next~\cite{li2024llavainterleave}, VILA~\cite{lin2023vila}, InternVL~\cite{chen2024internvl}), and recent unified foundation models (GPT-5~\cite{openai2024gpt4ocard}, Gemini2.5~\cite{comanici2025gemini25}, Qwen3~\cite{yang2025qwen3technicalreport}). While these models excel in text generation and high-level reasoning, they lack inherent spatial localization capabilities. Recent works have integrated pixel-level localization through segmentation adapters (LISA~\cite{lai2024lisa}, VISA~\cite{yan2025visa}), region encoders (GLaMM~\cite{rasheed2024glamm}, Ferret~\cite{you2023ferret,zhang2024ferretv2improvedbaselinereferring}), and multi-prompt frameworks (SA2VA~\cite{yuan2025sa2va}). However, these VLMs operate solely on 2D images or videos and cannot process 3D volumetric medical data.

\subsection{Medical VLMs}
The integration of multimodal reasoning into medical imaging has led to rapid progress in medical VLMs for report generation and VQA. Early efforts such as LLaVA-Med~\cite{li2023llava-med} and Med-Flamingo~\cite{pmlr-v225-moor23a} adapt general-purpose multimodal frameworks to radiology datasets, enabling report generation and VQA. Subsequently, HuatuoGPT-Vision~\cite{zhang2023huatuogpttaminglanguagemodel}, HealthGPT~\cite{lin2025healthgptmedicallargevisionlanguage}, MedTrinity~\cite{xie2025medtrinity25mlargescalemultimodaldataset}, and RadFM~\cite{wu2025towards} extend this line of work by coupling visual-language reasoning with radiology-specific representations to improve clinical understanding. More recent models, including MIMO~\cite{chen2024mimo} and MedPLIB~\cite{huang2024biomedpixelllm}, improve fine-grained anatomy detection through language-guided segmentation. However, all these models are constrained to 2D slice-based analysis and cannot capture volumetric spatial continuity, which is essential for accurate 3D medical reasoning. To address this limitation, several models have advanced toward volumetric understanding. CT-CHAT~\cite{hamamci2024generalistCT} and Med3DVLM~\cite{xin2025med3dvlm} align 3D CT embeddings with radiology reports for volumetric text generation but lack segmentation capabilities. VividMed~\cite{luo2025vividmedvisionlanguagemodel}, MS-VLM~\cite{lee2024readlikeradiologistefficient} and Med-2E3~\cite{shi2025med2e32denhanced3dmedical} extend the CT-CHAT model by integrating both 3D and 2D features, but remain restricted to text-only tasks. 
M3D~\cite{bai2024m3d} introduces segmentation heads for spatial reasoning but suffers from information loss caused by the compression of input image dimension and projection layer, while VILA-M3~\cite{nath2024vilam3} employs an agent-triggered segmentation mechanism that highly depends on explicit token activation.

\subsection{Vision Foundation Models}
The Segment Anything Model (SAM)~\cite{kirillov2023segment} and SAM2~\cite{ravi2024sam2segmentimages} establish universal segmentation capabilities through multi-prompt interaction (points and boxes). Their strong zero-shot generalization abilities have motivated adaptations to the medical domain~\cite{CHEN2024103310, shaharabany2023autosamadaptingsammedical}: MedSAM~\cite{Ma_2024} and MedSAM2~\cite{ma2024segmentmedicalimagesvideos} fine-tune SAM architectures on radiology datasets, while SAM-Med3D~\cite{yang2023sam3dsegment3dscenes}, SegVol~\cite{du2024segvol}, and parameter-efficient adapters~\cite{wu2023medicalsamadapteradapting} extend these models to 3D volumetric data. Additional efforts such as UniSeg~\cite{butoi2023universeguniversalmedicalimage} and BiomedParse~\cite{zhao2025biomedparse} further broaden biomedical segmentation across multiple imaging modalities. Despite their strong perceptual capability, these VFMs remain segmentation-oriented and lack language-guided reasoning. Our framework addresses this gap by integrating SAM2’s promptable segmentation within a LLaVA-style 3D medical VLM backbone, combining precise mask generation with semantic understanding to enable unified 3D volumetric reasoning and interactive segmentation. A comparison of task coverage across existing SOTA 3D medical VLMs, VFMs, and our model is provided in Table.~\ref{tab:task_comparison}.

\section{Methods}
\subsection{Problem definition}
3D medical vision-language tasks cover a broad spectrum, including report generation, VQA, referring segmentation, and interactive segmentation. To support all these tasks within a single model, we introduce a unified formulation that accommodates both text-centric and mask-centric predictions. Given a 3D medical volume $\mathcal{I}_{in}\in\mathbb{R}^{X\times{Y}\times{Z}}$ with size of $X\times{Y}\times{Z}$, a sequence of text tokens $\mathcal{T}_{in}\in \mathbb{R}^{N\times{D}}$ with size of $N\times{D}$, and a set of prompt tokens $\mathcal{P}_{in}$, the unified multi-task problem can be formulated as
\begin{equation}
\left(\mathcal{T}_{out},\mathcal{M}_{out}\right) = \mathcal{F}\left(\Theta;\mathcal{I}_{in},\mathcal{T}_{in},\mathcal{P}_{in}\right),
\end{equation}
where $\mathcal{F}(\cdot)$ denotes the unified 3D medical VLM, $\Theta=\{\theta_{vlb}, \theta_{seg}\}$ denotes the trainable parameters including the vision-language backbone parameters $\theta_{vlb}$ and the segmentation module parameters $\theta_{seg}$, $\mathcal{T}_{out}$ represents the output text tokens, and $\mathcal{M}_{out}$ is the output segmentation mask. The output text tokens  $\mathcal{T}_{out}$ are decoded by the tokenizer to generate the text output. Different from the current unified medical VLMs, our method provides visual prompts $\mathcal{P}_{in}$ to enhance interactive segmentation. Here $\mathcal{P}_{in}$ includes the bounding box prompt $\mathcal{P}_{bbx}$ and the point prompt $\mathcal{P}_{pt}$.
\subsection{Architecture Overview}
\begin{figure*}
  \centering
  \includegraphics[width=0.94\linewidth]{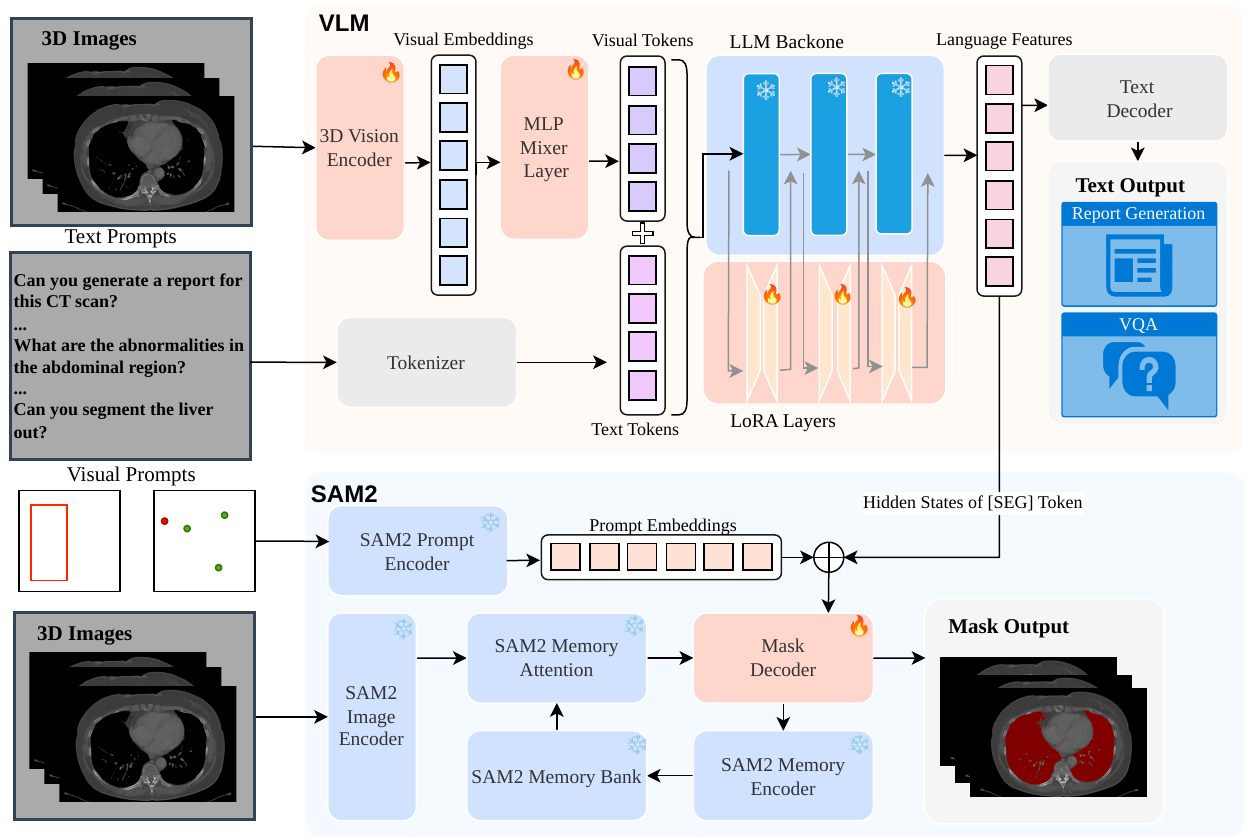}
  \caption{Overview of the proposed architecture, which integrates a LLaVA-style VLM with a SAM2-based segmentation module. The VLM processes 3D volumes and generates text autoregressively. When a \texttt{[SEG]} token is produced, its hidden state is extracted and fed into SAM2’s prompt encoder, where it is fused with optional visual prompts (points or boxes) to generate the final segmentation mask.}
  \label{fig:architecture}
\end{figure*}

As shown in Fig.~\ref{fig:architecture}, our approach consists of a LLaVA-like VLM and a promptable segmentation module. The VLM comprises the following components: (i) a volumetric vision encoder, (ii) an adaptive token projection layer, and (iii) an LLM backbone with LoRA layers. Our architecture addresses three key challenges in unified 3D medical VLM design: 
\textbf{Challenge 1: Volumetric Understanding.} Unlike 2D VLMs, 3D medical imaging requires capturing spatial relationships across slices. We address this through a 3D-aware vision encoder (M3D-CLIP) that preserves volumetric context.
\textbf{Challenge 2: Computational Efficiency.} Direct encoding of 3D volumes produces 2048 visual tokens per image (fourfold more than 2D LLaVA), causing prohibitive LLM inference costs. We address this via a MLP-Mixer projection layer that reduces tokens to 512 while preserving spatial information.
\textbf{Challenge 3: Cross-Slice Consistency.} Unlike 2D segmentation, where each image is independent, 3D segmentation requires anatomical continuity across slices. To alleviate this issue, the proposed model leverages the memory attention mechanism of SAM2 to maintain slice-wise consistency throughout the volume.
\subsection{Vision-Language Processing Pipeline}
\noindent\textbf{Volumetric vision encoder}. Following the design of LLaVA, a CLIP-based 3D encoder~\cite{radford2021learningtransferablevisualmodels} is adopted as the vision encoder. Given an input 3D volume $\mathcal{I}_{in}$, the image embeddings $v_{in}$ are derived via a 3D CLIP-based encoder $E_{vc}$ resulting in $v_{in}=E_{vc}(\mathcal{I}_{in})\in\mathbb{R}^{n \times d}$, where $n$ is the number of visual tokens and $d$ is the token dimension. In our framework, model weights from the pre-trained 3D vision-transformer (3D ViT)~\cite{bai2024m3d} are utilized.

\noindent\textbf{Adaptive token projection layer}. The high spatial dimensionality of 3D medical images leads to very large token sequences after ViT patch embedding. In 3D CLIP, a $32 \times 256 \times 256$ volume is partitioned into $8 \times 16 \times 16$ patches, producing $2048$ visual tokens, which results in a four-fold increase in inference time compared with the $512$-token LLaVA model. Furthermore, because anatomical structures occupy only a small portion of the volume, most visual tokens correspond to background regions, making direct processing of all tokens highly inefficient.

To address this issue, we introduce two MLP-Mixer layers~\cite{tolstikhin2021mlpmixerallmlparchitecturevision} as a projector that compresses the visual tokens from 2,048 to 512 with minimal information loss. Each MLP-Mixer layer contains two fully connected sublayers with nonlinearities applied independently to each row of the input tensor: the first performs channel mixing, and the second performs token mixing. Omitting layer indices, the mixer layer is formulated as
\begin{equation}
    \begin{split}
    U^T =  W_2\sigma(W_1\ \text{LayerNorm}(v_{in}^T)),\\
    T_v = W_4\sigma(W_3\ \text{LayerNorm}(U)),\\
    \end{split}
    \label{eq:mixer}
\end{equation}
where $\sigma$ denotes the GELU activation function, $U\in\mathbb{R}^{\hat{n}\times d}$ and $T_v \in \mathbb{R}^{\hat{n} \times \hat{d}}$ are the visual tokens aligned with the LLM's text embedding dimensions ( $\hat{n}=512$ and $\hat{d}=2048$ in our case). The computational complexity of the MLP-Mixer layer is linear with respect to the number of input patches, which enables efficient compression of 3D volumetric embeddings while preserving spatial information.

\noindent\textbf{LLM backbone with LoRA}. In our model, InternVL-2.5-4B~\cite{chen2024internvl} is utilized as the LLM backbone for its strong multimodal understanding and text generation capabilities. LoRA~\cite{hu2021loralowrankadaptationlarge} layers are applied to the query, key, and value projection matrices in all transformer layers to enable efficient fine-tuning. The overall process is formulated as: $T_{out} = LLM(T_{in}+T_v)$, where $T_{{in}}$ denotes the tokenized text prompt and $T_v$ denotes the visual tokens produced by the projection layer.
\subsection{\texttt{[SEG]} Token Controlled Segmentation}
In our framework, SAM2 is employed as the segmentation module for its strong performance and native support for multi-prompt interaction. The 3D medical volume is processed slice-by-slice using memory attention to maintain cross-slice consistency. Voxel-level segmentation is enabled through a hidden state of \texttt{[SEG]} token, denoted as \texttt{[SEG]}$_{hs}$, which serves as a semantic prompt and complements geometric prompts such as points and bounding boxes. Specifically, \texttt{[SEG]}$_{hs}$ is passed to the SAM2 decoder, where it is concatenated with prompt embeddings from the SAM2 prompt encoder and decoded into segmentation masks. During training, the SAM2 decoder learns to interpret the spatial-temporal prompt, while gradients flowing through the \texttt{[SEG]} token enable the VLM to learn how to use it for accurate localization. The segmentation module can be denoted as $\mathcal{M}_{out} = SAM2(\mathcal{I}_{in}, \texttt{[SEG]}_{hs}, \mathcal{P}_{in})$. A unidirectional design is adopted in which SAM2 outputs are not fed back into the LLM, allowing the architecture to remain simple, minimizing cross-modal alignment overhead, and enabling independent module upgrades.

\section{Implementation Details}
\subsection{Datasets}
\noindent\textbf{CT-RATE}. To evaluate  report generation and VQA performance, we utilize the CT-RATE dataset~\cite{hamamci2024generalistCT}, a large-scale, public resource comprising $21,304$ 3D non-contrast chest CT volumes paired with corresponding free-text radiology reports and over 2.7 million VQA pairs. The dataset is officially partitioned into a training set of $20,000$ patients and a test set of $1,304$ patients. For report generation, the “Findings” and “Impression” sections are extracted from the original radiology reports and used as ground truth. A comprehensive instruction-tuning dataset derived from the CT-RATE reports is employed for VQA, which is specifically structured to cover three distinct question-answer formats, including short answers, long answers, and multiple-choice questions, to evaluate the robustness of the model's interactive and diagnostic reasoning capabilities. 
\noindent\textbf{M3D-Seg}. The M3D-Seg dataset~\cite{bai2024m3d} is used to evaluate 3D segmentation performance. M3D-Seg is a unified benchmark that aggregates 3D CT scans from 25 public segmentation datasets, providing 5,772 labeled volumes paired with 119,267 category text descriptions and segmentation masks covering 211 anatomical structures. 
For interactive segmentation, we generate point and bounding-box prompts following the MedSAM2 protocol: three positive points are randomly sampled within the target region, and boxes are derived from the ground-truth masks with $\pm5\%$ spatial jitter to simulate imperfect user input. 
\begin{table}[t]
  \caption{Module parameters and model size}
  \label{tab:parameter_size}
  \centering
  \small
  \begin{tabular}{lcc}
    \toprule
    Module & Parameters & Size \\
    \midrule
    Vision Encoder & 198.05M  & 377.76 MB \\
    Projection Layer & 7.09M  & 13.52 MB \\
    LLM Backbone & 3.39B  & 6.33 GB \\
    Segmentation Module & 224.43M  & 428.07 MB \\
    \midrule
    Total & 3.82B & 7.15 GB \\
    \bottomrule
  \end{tabular}
\end{table}
\begin{figure*}[t]
  \centering
  \includegraphics[width=\linewidth]{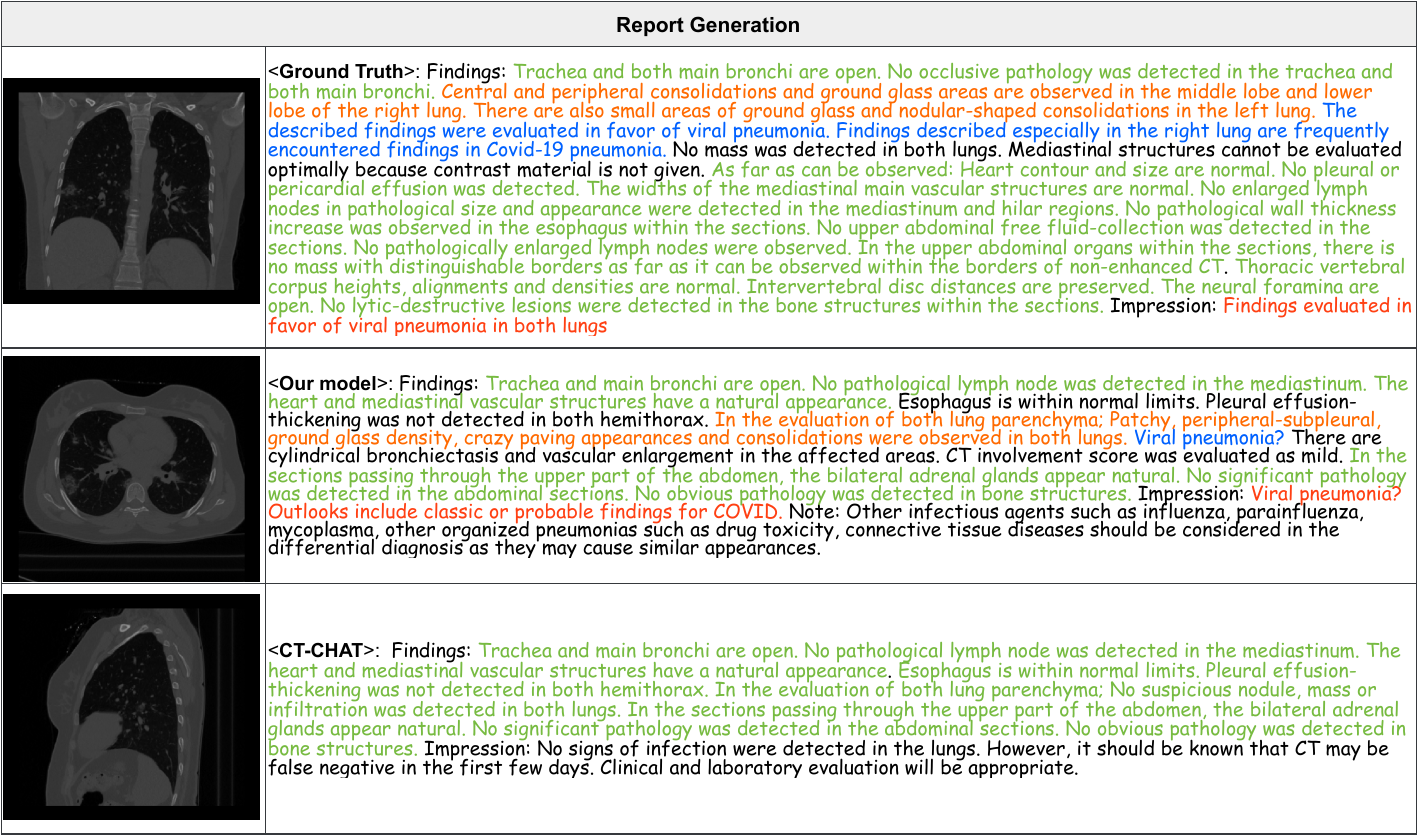}
  \caption{Qualitative comparison of the proposed method with CT-CHAT on 
  the report generation task. Coronal, axial, and sagittal views of 
  the corresponding CT scans are shown. Different aspects of the reports are 
  color-coded for improved readability.}
  \label{fig:report_generation}
\end{figure*}
\subsection{Multi-Stage Training}
\begin{figure*}[t]
  \centering
   \includegraphics[width=\linewidth]{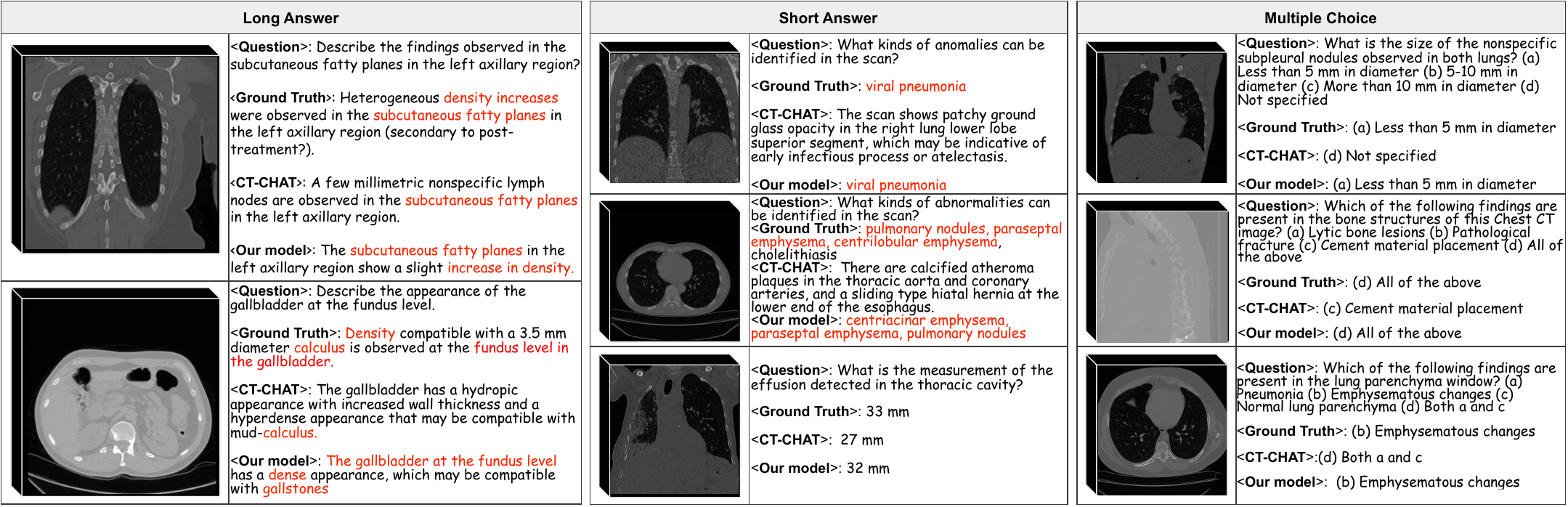}

   \caption{Comparison between the proposed methods to CT-CHAT on VQA tasks, including Long Answer, Short Answer, and Multiple choice subsets. Critical clinical information was highlighted in red. For CT-CHAT, the input question additionally contained a special token (\texttt{[short\_answer]}, \texttt{[long\_answer]}, \texttt{[multiple\_choice]}) to indicate the specific subset to evaluate.}
   \label{fig:text_task}
\end{figure*}

Inspired by Cambrian-1~\cite{tong2024cambrian1} framework, we employ a three-stage progressive training strategy to stably align 3D visual features, language representations, and voxel-level localization capability. In the first stage, only the MLP-Mixer projection layer is optimized, while the vision encoder, LLM, and SAM2 modules remain frozen. This stage establishes the initial vision-language alignment and is trained for 3 epochs on the report-generation subset of CT-RATE using the autoregressive language modeling loss $\mathcal{L}_{text}$~\cite{yuan2025sa2va}. In the second stage, the trainable set is expanded to include the vision encoder, the projection layer, and the LLM LoRA parameters, with SAM2 still frozen. This stage focuses on strengthening cross-modal representation learning and is trained for 5 epochs on CT-RATE, sampling report-generation, and VQA tasks with equal probability under the same objective $\mathcal{L}_{text}$. In the final stage, the SAM2 decoder is unfrozen to enable full multimodal spatial localization and segmentation. Joint optimization is performed for 5 epochs using both the CT-RATE and the M3D-Seg datasets using the composite objective function
\begin{equation}
\mathcal{L}_{joint}
=\lambda_{text}\mathcal{L}_{text}
+\lambda_{mask}\left(\mathcal{L}_{CE}
+\lambda_{Dice}\mathcal{L}_{Dice}\right),
\end{equation}
where $\mathcal{L}_{CE}$ is the voxel-wise binary cross-entropy loss, $\mathcal{L}_{Dice}$ denotes the Dice loss. Coefficients $\lambda_{text}$, $\lambda_{mask}$, and $\lambda_{Dice}$ control the relative contributions of each term. This staged training schedule reduces catastrophic forgetting, stabilizes optimization, and results in a unified model capable of robust image-text alignment and mask prediction. Detailed module parameters of the proposed method are presented in 
Table~\ref{tab:parameter_size}. Eight NVIDIA B200 GPUs were used for training 
and one B200 GPU was used for evaluation. For training Stages 1 and 2, 
where the trainable parameters did not include SAM2 modules, the batch 
size was set to 16 per GPU (effective batch size 128). Training Stage 1 
for vision-language alignment required approximately 6 hours over 3 
epochs, while training Stage 2, which strengthened language reasoning, 
required approximately 18 hours over 5 epochs. For training Stage 3 
involving joint-training, the batch size was set to 1 per GPU with 
8-step gradient accumulation (effective batch size 8). The entire 
joint-training process required approximately 110 hours over 10 epochs.
Loss function hyperparameters for Stage 3 were set as follows: 
$\lambda_{text} = 0.5$, $\lambda_{mask} = 2.0$, and $\lambda_{Dice} = 1.0$. 
These values were determined through validation set tuning.
\section{Experiments}
\begin{table}
  \caption{Performance comparison on report generation.}
  \label{tab:report_generation}
  \centering
  \resizebox{\columnwidth}{!}{
  \begin{tabular}{lccccc}
    \toprule
    Method & BLEU-1 &ROUGE & METEOR & CIDEr  & BERTScore\\
    \midrule
    LLaVA~\cite{liu2023vit}      & 18.47 & 10.34 & 11.22 & 0.050 & 30.55 \\
    CXR-LLaVA~\cite{lee2024cxrllavamultimodallargelanguage}  & 9.12  & 5.05  & 2.86  & 0.049 & 39.18 \\
    LLaVA-Med~\cite{li2023llava-med}  & 9.01  & 5.24  & 2.24  & 0.006 & 40.65 \\
    M3D~\cite{bai2024m3d}        & 14.49 & 19.25 & 14.11 & 0.081& 84.12 \\
    CT-CHAT~\cite{hamamci2024generalistCT}    & \textbf{43.64} & 31.57 & \textbf{23.06} & 0.221& 88.12 \\
    Ours       & 41.85 & \textbf{34.59} & 22.15 & \textbf{0.237}& \textbf{89.33} \\
    \bottomrule
  \end{tabular}
  }
\end{table}
\subsection{Evaluation on Report Generation}
\begin{figure*}[t]
  \centering
  \includegraphics[width=\linewidth]{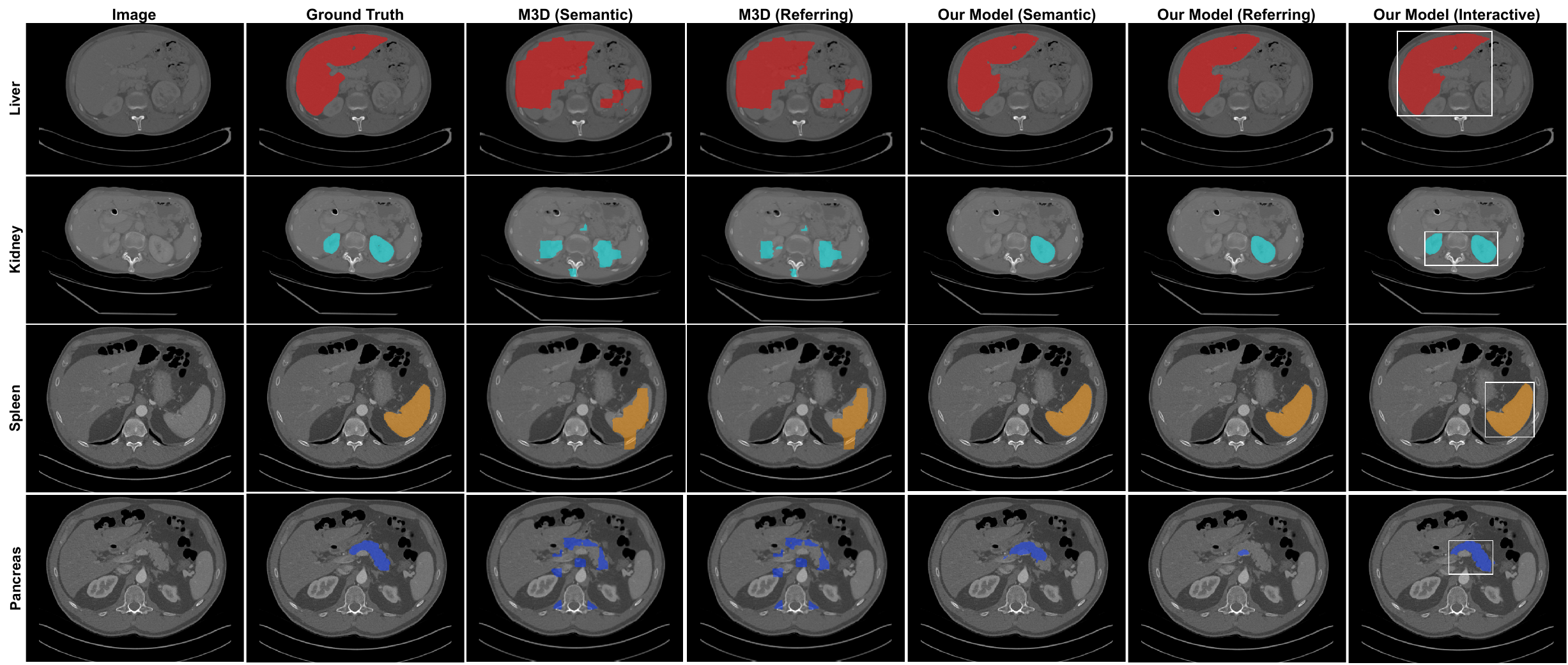}
  \caption{Qualitative comparison of the proposed method with M3D on 
  referring and semantic segmentation. Interactive segmentation results using 
  bounding-box prompts are also included for comparison (ACT-1K dataset). 
  The liver is shown in red, the kidney in cyan, the spleen in 
  orange, and the pancreas in blue.}
  \label{fig:seg_task_act}
\end{figure*}
We evaluate report generation on the CT-RATE test set, which contains 1,304 CT-report pairs. Traditional n-gram-based NLP metrics and LLM-based metrics are used for assessment, including BLEU~\cite{10.3115/1073083.1073135}, ROUGE~\cite{lin-2004-rouge}, METEOR~\cite{10.5555/1626355.1626389}, CIDEr~\cite{vedantam2015cider}, and BERTScore~\cite{zhang2020bertscoreevaluatingtextgeneration}. While n-gram metrics capture lexical overlap, they offer limited semantic insight. In contrast, LLM-based metrics assess a deeper semantic similarity between the generated reports and the ground-truth findings. We compare our model with representative multimodal baselines, including LLaVA~\cite{liu2023vit}, CXR-LLaVA~\cite{lee2024cxrllavamultimodallargelanguage}, LLaVA-Med~\cite{li2023llava-med}, M3D~\cite{bai2024m3d}, and CT-CHAT~\cite{hamamci2024generalistCT}. All models are evaluated under identical inference conditions. The results are presented in Table.~\ref{tab:report_generation} and qualitative examples are shown in Fig.~\ref{fig:report_generation}. Different aspects of the generated reports are color-coded for clarity: clinical findings in orange, COVID-19 related findings in blue, normal/healthy findings in green, and diagnostic conclusions in red. Our model achieves strong performance across both traditional and LLM-based evaluation metrics. Specifically, it outperforms 2D medical multimodal models such as LLaVA, CXR-LLaVA, and LLaVA-Med by large margins, indicating better textual fluency and clinical content accuracy. Compared with 3D baselines, our model significantly surpasses M3D and achieves comparable or higher scores than the current state-of-the-art CT-CHAT, demonstrating improved semantic fidelity and reasoning ability in the volumetric report generation task. Task-specific templates were used for report generation task. Representative templates are provided in Appendix A.
\subsection{Evaluation on VQA}
\begin{figure*}[t]
  \centering
   \includegraphics[width=\linewidth]{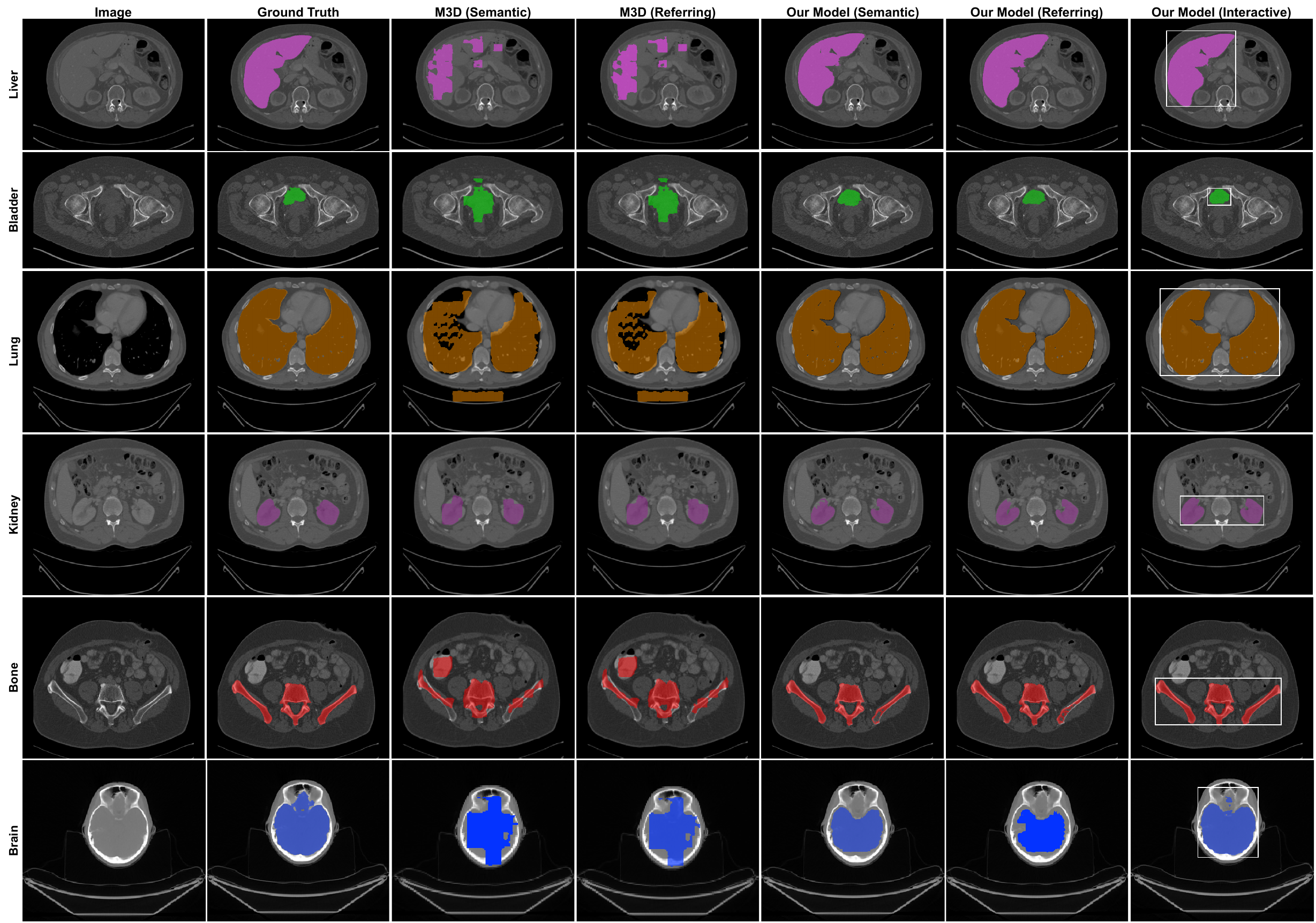}
   \caption{Comparison of the proposed method with M3D on referring and semantic segmentation. Interactive segmentation results using bounding-box prompts are also included for comparison (CTOrg dataset shown). Liver is shown in magenta, bladder in green, lung in brown, kidney in purple, bone in red, and brain in blue.}
   \label{fig:seg_task}
\end{figure*}
We evaluated VQA performance on the CT-RATE dataset, which includes three subtasks: long-answer, short-answer, and multiple-choice VQA. Qualitative examples are shown in Fig.~\ref{fig:text_task}. The long-answer and short-answer subtasks are treated as open-ended VQA, where the model generates free-form responses, while the multiple-choice subtask is treated as closed-ended VQA, where the model selects one answer from predefined options. For open-ended tasks, we employ BLEU, ROUGE, METEOR, CIDEr, and LLM score to evaluate lexical and semantic similarity between generated and reference answers. For the closed-ended task, accuracy is used as the measure. Based on the findings from the report-generation experiments, which reveal that 2D medical multimodal models such as LLaVA, CXR-LLaVA, and LLaVA-Med, and the 3D multimodal model M3D exhibit substantial performance limitations, we focus our comparison on the current state-of-the-art CT-CHAT model. Quantitative results are presented in Table.~\ref{tab:VQA}. Our model achieves comparable or superior performance across all VQA subtasks. It delivers consistent improvements in ROUGE, LLM score, and accuracy, indicating stronger semantic alignment and reasoning capabilities. Notably, our model achieves 89.74\% accuracy on multiple-choice VQA, surpassing CT-CHAT by a clear margin. These results confirm that incorporating prompt-driven segmentation and volumetric understanding enhances multimodal reasoning and overall comprehension in both open-ended and closed-ended 3D medical VQA tasks.

\begin{table*}[t]
  \centering
  \caption{Comparison between CT-CHAT and our model on long-answer, short-answer, and multiple-choice VQA tasks.}
  \label{tab:VQA}
  \small
  \setlength{\tabcolsep}{4pt}
   \resizebox{\textwidth}{!}{
  \begin{tabular}{l|ccccc|ccccc|c}
    \toprule
    \multirow{2}{*}{Task} & \multicolumn{5}{c|}{Long Answer} & \multicolumn{5}{c|}{Short Answer} & \multicolumn{1}{c}{Multiple Choice} \\
    \cmidrule(lr){2-6} \cmidrule(lr){7-11} \cmidrule(lr){12-12}
    & BLEU & ROUGE & METEOR & CIDEr & BERTScore & BLEU & ROUGE & METEOR & CIDEr & BERTScore & Accuracy \\
    \midrule
    CT-CHAT~\cite{hamamci2024generalistCT} & \textbf{47.020} & 48.450 & 28.200 & \textbf{2.910} & 90.140 & \textbf{27.460} & 45.590 & \textbf{15.470} & 1.652 & 90.120 & 83.730 \\
    Ours & 45.941 & \textbf{48.983} & \textbf{28.218} & 2.861 & \textbf{92.760} & 24.076 & \textbf{57.428} & 14.026 & \textbf{1.771} & \textbf{92.503} & \textbf{89.741} \\
    \bottomrule
  \end{tabular}
  }
\end{table*}
\subsection{Evaluation on Referring Segmentation}
We evaluated referring segmentation on the M3D-Seg test set using the same data split as the M3D model. Since no other existing models support referring segmentation for 3D medical volumes, we use only SegVol and M3D as baseline methods. Performance is reported on three benchmark subsets of M3D-Seg dataset, including CTOrg~\cite{rister2020ctorg}, Abdomen-CT-1K (ACT-1K)~\cite{9497733}, and TotalSegmentator (TS)~\cite{doi:10.1148/ryai.230024}, using the Dice coefficient as the evaluation metric. For referring segmentation, the text prompt is a randomly selected free-form description of the target region that excludes explicit class names, allowing us to evaluate the model’s ability to perform language-driven spatial reasoning. To further assess the reasoning capabilities of our model, we additionally evaluate performance on semantic segmentation, where the text prompt explicitly specifies the anatomical class (e.g., “segment the liver”). Task-specific templates were used for referring segmentation and semantic segmentation tasks. Representative templates are provided in Appendix A.

Quantitative results are shown in Table.~\ref{tab:refer_seg}. Across all subsets, our model achieves the highest Dice scores in both referring and semantic segmentation. Compared to SegVol, which relies solely on volumetric features, our model improves Dice by at least 10 percent for each dataset. When compared to M3D, our approach achieves consistent gains of around 5 percent, demonstrating stronger spatial reasoning and text-conditioned localization abilities. Notably, our referring segmentation performance is comparable to semantic segmentation, indicating that the model can accurately localize anatomical structures from implicit textual descriptions without explicit class labels. Overall, these results validate the effectiveness of our unified 3D multimodal framework for language-guided volumetric segmentation. The integration of prompt-driven SAM2 segmentation with multimodal reasoning enables precise and interpretable 3D spatial segmentation capability for clinical workflows involving complex 3D anatomical structures.

\subsection{Evaluation on Interactive Segmentation}
We further evaluate interactive segmentation on the same dataset and testing protocol as in the referring segmentation experiments. This setting simulates a human-in-the-loop workflow, in which users iteratively refine segmentation masks through additional prompts. Two interaction modes are examined: bounding-box prompting and point prompting. 
For bounding-box prompting, a random slice is selected from each 3D volume, and a bounding box with slight spatial jitter from the ground truth is provided as input. For point prompting, three randomly sampled positive points within the target region are selected on a random slice. In both settings, the model receives the same image and text inputs as in referring segmentation, augmented with the corresponding visual prompts. Dice coefficient is used as the evaluation metric. Since no existing 3D medical VLMs support interactive segmentation with text-based prompts, our semantic and referring segmentation results serve as baselines. 

Quantitative results are summarized in Table.~\ref{tab:interactive_seg}, and qualitative examples for both referring and interactive segmentation are shown in Fig.~\ref{fig:seg_task_act} and  Fig.~\ref{fig:seg_task}. Across all datasets, both interactive variants consistently outperform their non-interactive baselines. Bounding-box prompts yield the highest Dice scores, followed by point prompts. These findings demonstrate that incorporating user feedback substantially improves segmentation quality.
We also observe that interactive refinement yielded modest improvements on CTOrg and ACT-1K. Both datasets consist primarily of large, well-defined organs (e.g., liver, lungs, kidneys), for which the language-driven \texttt{[SEG]} token already encodes rich spatial and semantic cues from the volumetric context. As a result, adding point or bounding-box prompts provides only incremental gains, since the model can infer coarse boundaries directly from language descriptions. In contrast, results on the TotalSegmentator dataset exhibit a larger improvement. This dataset is considerably more challenging, as it contains numerous small and complex anatomical structures, such as vertebrae, ribs, and vessels. For these fine-grained regions, the global language representation may lack sufficient spatial specificity, leading to under- or over-segmentation when relying solely on text.  In these cases, explicit interactive prompts supply critical spatial priors, enabling the model to localize small anatomical regions more precisely. Overall, these findings highlight the complementary role of interactive segmentation in clinical workflows: while large structures can often be segmented reliably from language alone, small or complex anatomical targets benefit significantly from user-guided refinement, making the model more practical and adaptable for real-world clinical workflows.

\begin{table}
  \caption{Results on the referring segmentation task.}
  \label{tab:refer_seg}
  \centering
  \small
  \begin{tabular}{lccccc}
    \toprule
    
    Method & CTOrg & ACT-1K & TotalSegmentator\\
    \midrule
    SegVol~\cite{du2024segvol}           & 77.78  & 79.06  & 44.28\\
    M3D~\cite{bai2024m3d} (semantic)   & 81.27  & 73.64  & 58.30\\
    M3D~\cite{bai2024m3d} (referring)  & 83.49  & 73.63  & 58.50 \\
    Ours (semantic)  & \textbf{88.04}  & 88.27  & 66.23 \\
    Ours (referring) & 87.93  & \textbf{88.34}  & \textbf{66.50}\\
    \bottomrule
  \end{tabular}
\end{table}
\begin{table}
  \caption{Results on the interactive segmentation task.}
  \label{tab:interactive_seg}
  \centering
  \small
  \begin{tabular}{lccccc}
    \toprule
    Method & CTOrg & ACT-1K & TotalSegmentator\\
    \midrule
    Ours (semantic)  & 88.04  & 88.27  & 66.23 \\
    Ours (referring) & 87.93  & 88.34  & 66.50\\
    Ours (bbox prompt)  & \textbf{88.76}  & \textbf{88.71}  & \textbf{70.51} \\
    Ours (points prompt) & 88.52  & 88.49  & 69.88\\
    \bottomrule
  \end{tabular}
\end{table}

\section{Ablation Studies}
\subsection{Impact of Vision Encoder and Projection Layer}
To assess the contributions of the vision encoder and projection layer, we conduct ablation experiments on the CTOrg and ACT-1K subsets using the interactive segmentation task. As shown in Table.~\ref{tab:ablation_components}, two vision encoders are compared: vanilla CLIP and M3D-CLIP. For the projection layer, we evaluate three configurations: Linear, MLP, and MLP-Mixer.
Replacing the vanilla CLIP encoder with the M3D CLIP backbone results in a substantial improvement in Dice scores, highlighting the importance of volumetric-aware pretraining for 3D medical data. The M3D CLIP encoder effectively captures cross-slice spatial context and anatomical continuity, while the vanilla CLIP model trained on natural images lacks such volumetric priors. Projection-layer design also has a notable impact. Linear and MLP projectors offer limited cross-modal alignment, resulting in weaker performance. In contrast, the MLP-Mixer achieves the highest Dice scores, indicating that its ability to jointly model spatial and channel-wise interactions enables richer multimodal fusion and more effective volumetric reasoning. Overall, these findings validate our architectural choices: M3D-CLIP provides strong 3D semantic representations, and the MLP-Mixer projection layer facilitates robust vision-language alignment, jointly leading to significant gains in both referring and interactive segmentation tasks.

\begin{table}
  \caption{Ablation studies on different choices of vision encoders and projection layers.}
  \label{tab:ablation_components}
  \centering
  \small
  \begin{tabular}{lccccc}
    \toprule
    Vision Encoder & Projection Layer & CTOrg & ACT-1K\\
    \midrule
    Vanilla CLIP~\cite{radford2021learningtransferablevisualmodels}  &Linear     & 30.18  & 44.65 \\
    M3D-CLIP~\cite{bai2024m3d}     &Linear     & 61.99  & 74.58 \\
    M3D-CLIP~\cite{bai2024m3d}     &MLP        & 77.78  & 79.83 \\
    M3D-CLIP~\cite{bai2024m3d}     &MLP-Mixer & \textbf{88.04}  & \textbf{88.27} \\
    \bottomrule
  \end{tabular}
\end{table}
\subsection{Impact of Multi-Stage Training Strategy}
To evaluate the effectiveness of our multi-stage training approach, we compare it with an alternative one-stage end-to-end training strategy, where all components are trained simultaneously from scratch. Both strategies are evaluated on the CT-RATE dataset for the report generation task, using BLEU-1, ROUGE, METEOR, and CIDEr as evaluation metrics. Quantitative results are shown in Table.~\ref{tab:ablation_training}.
The multi-stage training strategy consistently outperforms one-stage end-to-end training across all metrics. The one-stage approach exhibits suboptimal convergence, as the simultaneous optimization of multiple large modules under competing cross-modal objectives hinders stable alignment between visual and language representations.  In contrast, the progressive training schedule provides clear optimization phases, facilitating more reliable vision-language alignment. These results validate the use of a progressive multi-stage training pipeline.  

\begin{table}
  \caption{Ablation studies on different choices of training strategies.}
  \label{tab:ablation_training}
  \centering
  \small
   \resizebox{\columnwidth}{!}{
  \begin{tabular}{lccccc}
    \toprule
    Training strategy & BLEU-1 & ROUGE & METEOR & CIDEr\\
    \midrule
    One-stage  & 38.58   & 32.53  & 20.89 & 0.2095\\
    Multi-stage  & \textbf{41.85}   & \textbf{34.59}  & \textbf{22.15} & \textbf{0.2371}\\
    \bottomrule
  \end{tabular}
  }
\end{table}
\section{Discussion}
The primary contribution of this work is the integration of image-level semantic reasoning with fine-grained volumetric perception within a unified 3D VLM. The proposed framework is capable of processing 3D medical images and supporting multiple tasks, including report generation, VQA, and various forms of 3D image segmentation. This is achieved through the integration of a prompt-driven segmentation module, a 3D vision encoder, a multimodal projector, and an LLM backbone. The unified design enables joint reasoning over volumetric image features and language, while providing spatially grounded outputs. Experimental results on the CT-RATE and M3D-Seg benchmarks indicate that the proposed framework achieves competitive performance in report generation and VQA tasks. In addition, it supports referring and interactive segmentation capabilities that are not available in existing text-centric 3D VLMs. These results suggest that incorporating segmentation-aware components into 3D VLMs may offer a practical pathway toward more interpretable and clinically useful multimodal systems.

Despite these advancements, the proposed framework has several limitations. First, the current architecture adopts a unidirectional design in which the outputs of the segmentation module (SAM2) are not fed back into the LLM. While this decoupled design simplifies implementation and enables independent optimization of each component, it limits the LLM's ability to leverage segmentation outputs for iterative reasoning or self-correction. Second, the evaluation is currently restricted to 3D CT data with a distributional discrepancy across tasks. Specifically, text-driven tasks are primarily trained and evaluated on chest CT datasets~\cite{hamamci2024generalistCT}, whereas segmentation tasks rely on whole-body CT datasets~\cite{bai2024m3d}. This anatomical mismatch may introduce domain shifts that affect model stability and generalization across different body regions. Future work will focus on incorporating bidirectional interaction mechanisms between segmentation and language components, as well as further evaluating and improving the robustness and generalizability of the proposed framework.

\section{Conclusion}
In this work, we introduce MedVL-SAM2, a unified 3D VLM that integrates SAM2’s promptable segmentation with a LLaVA-style multimodal framework. The model jointly supports report generation, VQA, and semantic, referring, and interactive segmentation, bridging image-level understanding with voxel level localization and segmentation. Experiments demonstrate promising performance across multiple 3D benchmarks, confirming MedVL-SAM2’s strengths in volumetric reasoning, fine-grained spatial localization, and interactive segmentation. 
\section*{Appendix A. Instruction Templates Used for LLM Training}
\begin{figure}[!htbp]
  \centering
  \includegraphics[width=0.95\linewidth]{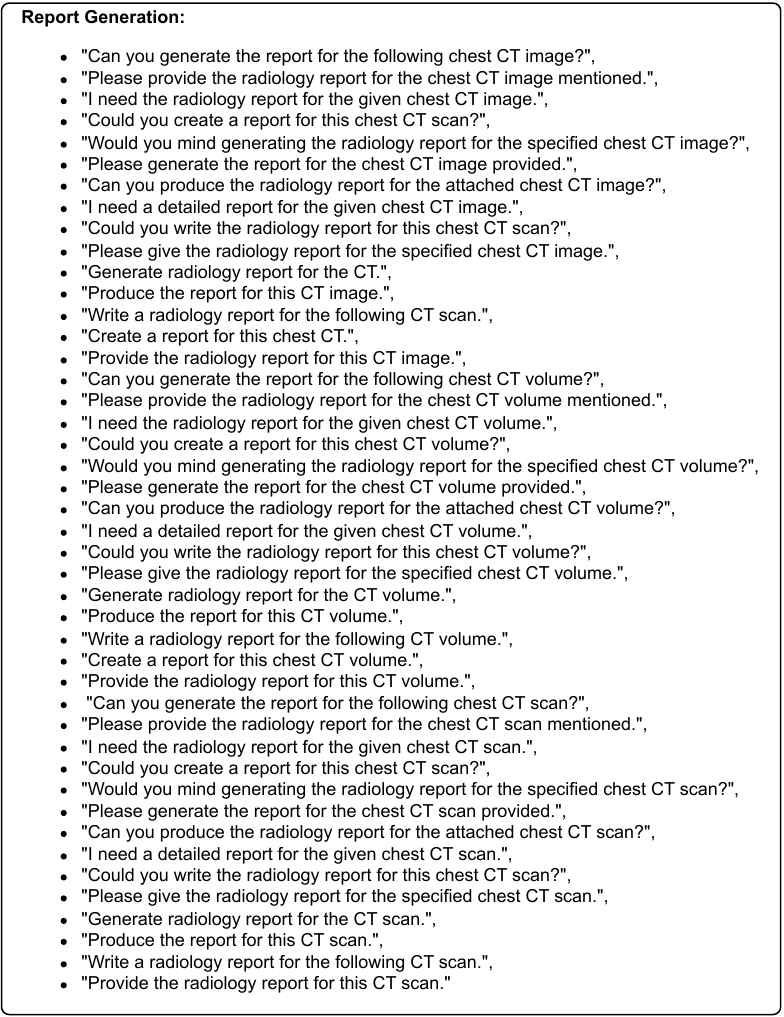}
  \caption{Prompt templates used for report generation tasks.}
  \label{fig:report_generation_template}
\end{figure}

\begin{figure}[!htbp]
  \centering
  \includegraphics[width=0.85\linewidth]{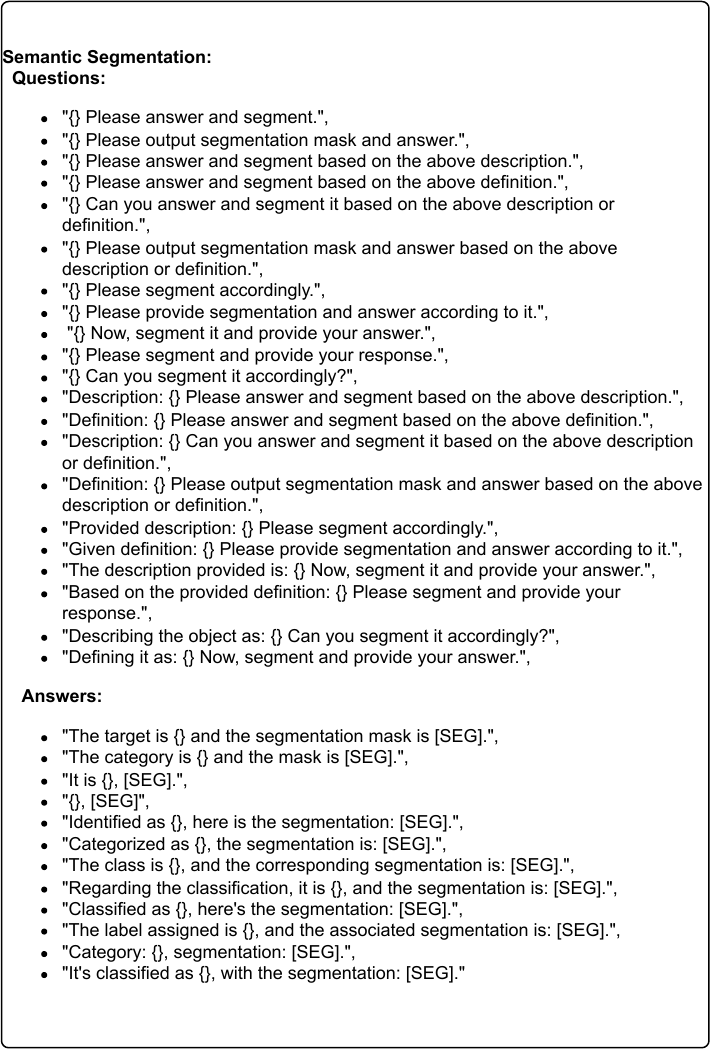}
  \caption{Prompt templates used for referring segmentation tasks.}
  \label{fig:rs_template}
\end{figure}

\begin{figure}[!h]
  \centering
  \includegraphics[width=0.83\linewidth]{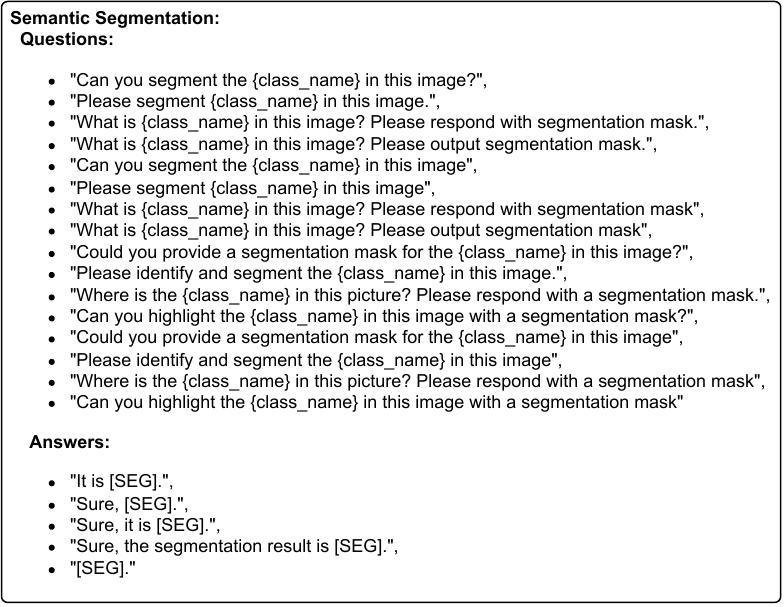}
  \caption{Prompt templates used for semantic segmentation tasks.}
  \label{fig:ss_template}
\end{figure}

\begin{figure}[!h]
  \centering
  \includegraphics[width=0.83\linewidth]{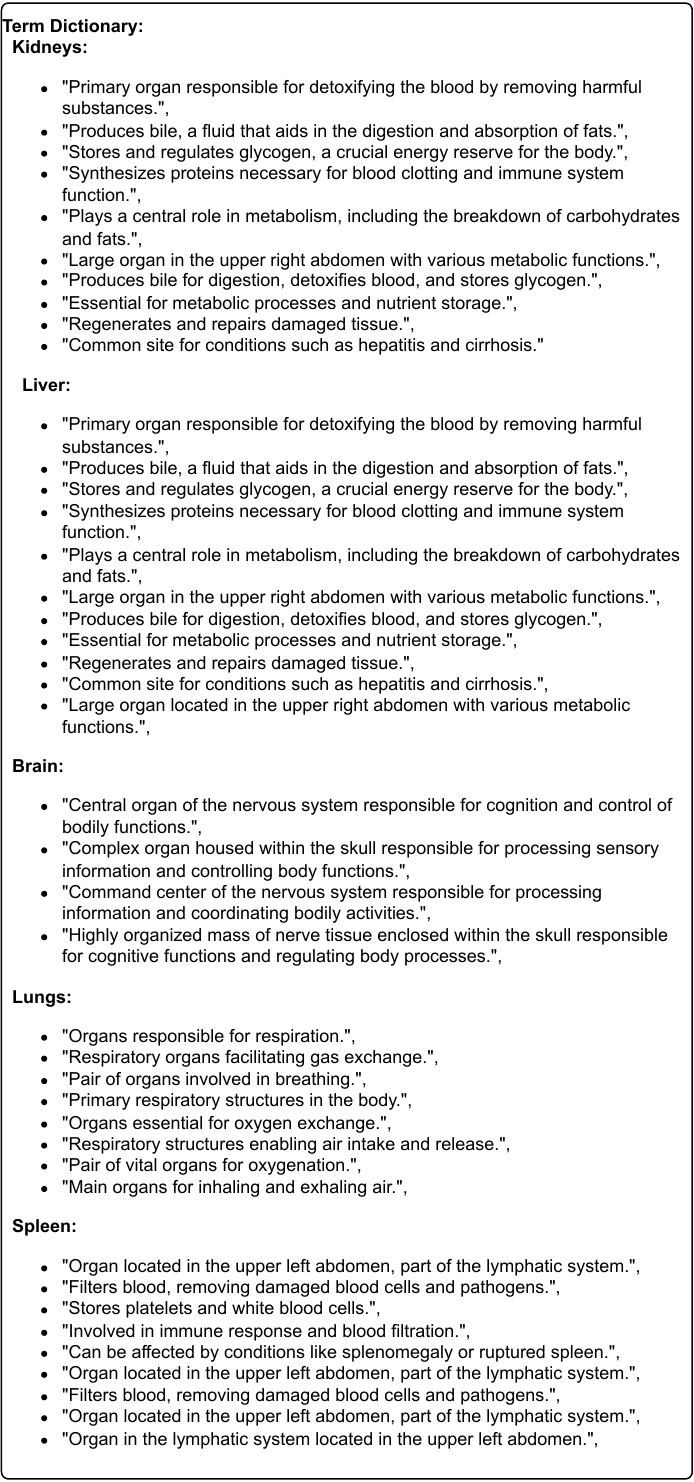}
  \caption{Anatomical descriptions used as free-form text prompts for 
  referring segmentation tasks. These descriptions replace the \{\} placeholder 
  in the templates shown in Figure~\ref{fig:rs_template}.}
  \label{fig:term_dict}
\end{figure}

This section demonstrates task-specific prompts and templates to guide the LLM across 
different tasks, including report generation, semantic segmentation, and 
referring segmentation. Figure~\ref{fig:report_generation_template} shows 
prompts for report generation, while Figures~\ref{fig:rs_template}, \ref{fig:ss_template}, and~\ref{fig:term_dict} present instruction templates 
for segmentation tasks.

\section*{References}
\bibliographystyle{IEEEtran}
\bibliography{main}

\end{document}